# Predictive Linear-Gaussian Models of Stochastic Dynamical Systems

**Matthew Rudary**  **Satinder Singh**  **David Wingate**

Computer Science and Engineering
University of Michigan
{mrudary,baveja,wingated}@umich.edu

## Abstract

Models of dynamical systems based on predictive state representations (PSRs) are defined strictly in terms of observable quantities, in contrast with traditional models (such as Hidden Markov Models) that use latent variables or state-space representations. In addition, PSRs have an effectively infinite memory, allowing them to model some systems that finite memory-based models cannot. Thus far, PSR models have primarily been developed for domains with discrete observations. Here, we develop the Predictive Linear-Gaussian (PLG) model, a class of PSR models for domains with continuous observations. We show that PLG models subsume Linear Dynamical System models (also called Kalman filter models or state-space models) while using fewer parameters. We also introduce an algorithm to estimate PLG parameters from data, and contrast it with standard Expectation Maximization (EM) algorithms used to estimate Kalman filter parameters. We show that our algorithm is a consistent estimation procedure and present preliminary empirical results suggesting that our algorithm outperforms EM, particularly as the model dimension increases.

## 1 INTRODUCTION

In many domains of interest, such as speech recognition, robot localization, and the stock market, we are interested in modeling a system that emits continuous observations. Linear Dynamical System models (LDSs), also called Kalman filter models or state-space models, are routinely used in control and prediction tasks in a wide variety of applications spanning many disciplines. While these models are very useful when their parameters are known in advance, these parameters are not easily *learned*, particularly as the model dimension increases. Because LDSs' state is unobserved (hidden), their parameters are typically learned using the Expectation Maximization (EM) algorithm. However, EM is only able to find parameters that maximize the data's likelihood locally, and may thus learn an inaccurate model.

In contrast, the state of a model based on *predictive state representations* (PSRs) (Littman, Sutton, & Singh, 2001) is a set of statistics defined over *future* observable outcomes.[1] PSRs have shown advantages over several traditional models: they are more expressive than HMMs (Jaeger, 1997; Singh, James, & Rudary, 2004), and learning PSRs from data may be more accurate than learning traditional models with hidden state (Wolfe, James, & Singh, 2005). An advantage of PSRs over traditional memory-based methods such as autoregressive and history window or $k^{th}$-order Markov models is that PSRs with a finite set of statistics about the future can model some domains that would require the memory-based models to use infinite memory. However, thus far PSR models are limited to domains with discrete observations.

We present a new PSR model class for continuous observations, the Predictive Linear-Gaussian model (PLG), and show that it subsumes classical LDS models (including Kalman filters and ARMA models), while using fewer parameters than an equivalent LDS of the same dimension. We present a learning algorithm that estimates PLG parameters from data traces, and prove that, under certain conditions, these estimators are *consistent*: As the number of traces increases, the estimated parameters converge in probability to their true values. This is in contrast to the local-optima guarantees of the EM algorithm for LDSs. Our algorithm's stronger guarantees crucially exploit the fact that a PLG's state uses statistics of observable quantities. We also present preliminary results suggesting that as the data set grows, our algorithm learns PLGs that outperform LDSs learned via EM, particularly as the dimension of the models increases.

---

[1] Jaeger (1997) and Rivest and Schapire (1994) also developed models based on predictions of future observable outcomes; PSRs have built on this earlier work.

## 2 THE PREDICTIVE LINEAR-GAUSSIAN MODEL

A model of a dynamical system allows one to make predictions about future events given any history. This is the modeling problem: to find and represent structure in the system in order to make predictions. We focus on modeling uncontrolled dynamical systems that emit scalar, real-valued observations at discrete time intervals.

One of the simplest models of this type is the $n$-dimensional Autoregressive (AR) model, which posits that the observation at any time step is a linear combination of the previous $n$ observations plus an independent noise term. Let $Y_t$ be the random variable modeling the observation at time step $t$ ($y_t$ will represent a realization of that variable). Then the AR model's dynamics at time $t$ is defined by

$$Y_t = a_t + \sum_{i=1}^{n} \phi_i y_{t-i}, \qquad (1)$$

where $a_t \sim \mathcal{N}(0, \sigma_{AR}^2)$ is the i.i.d. noise term and $\mathcal{N}(\mu, \Sigma)$ denotes the Gaussian distribution with mean $\mu$ and covariance matrix $\Sigma$.

A drawback to the AR model is that it has a finite memory: Given the history of observations up to time $t$, the observations up to time $t - n$ no longer affect the dynamics of the system. That is, if $B$ is any future event in the system, $\Pr(B|y_1, \ldots, y_t) = \Pr(B|y_{t-n+1}, \ldots, y_t)$.

The LDS model addresses the finite-memory limitation by adding hidden state to the mix: The (unobserved) state vector at a particular time is a noisy linear function of the previous state vector, and the observation is a noisy linear function of the state. Formally, let $X_t$ be the state vector at time $t$. The initial state is drawn from a normal distribution with mean $\widehat{x}_1^-$ and covariance $P_1^-$:

$$X_1 \sim \mathcal{N}(\widehat{x}_1^-, P_1^-). \qquad (2)$$

The observation at time $t$ is a linear function of the state, plus mean-zero i.i.d. noise:

$$Y_t | X_t = x_t \sim \mathcal{N}(Hx_t, R), \qquad (3)$$

where $H$ is the linear mapping from the state vector to the observation, and $R$ is the variance of the noise. Finally, the next state is a linear function of the current state, again with mean-zero i.i.d. noise added:

$$X_{t+1} | X_t = x_t \sim \mathcal{N}(Ax_t, Q), \qquad (4)$$

where $A$ is the linear trend in the state space and $Q$ is the covariance matrix of the noise. Because the state vector is unobservable, a distribution over underlying states must be computed; this is accomplished using the Kalman filter (see Appendix A.2).[2] The state vector can retain information about observations that are arbitrarily far in the past.

This infinite-memory property can be retained without positing unobservable state variables. Here, we develop such a model, the Predictive Linear-Gaussian model, and show that it is as powerful as the LDS model. A PLG model makes four assumptions about the structure of the system; as we state these assumptions, we develop the mathematics underlying the formal model.

Our *first assumption* is that, for some finite $n$, the distribution of all the future observations is completely characterized by the distribution of the next $n$ observations. That is, the distribution of the $n$ following observations serves as the state of the system; equivalently, this distribution serves as a *sufficient statistic* for history in this system. Note that this is *not* the $n^{th}$-order Markov assumption; state as full distributional information about the *next* $n$ observations is quite different than state as knowledge of the actual *past* $n$ observations.

Our *second assumption* is that all observations are jointly distributed according to a multivariate Gaussian distribution. This an extremely useful restriction. The Gaussian distribution has the conjugacy property; that is, if a set of variables $X_1, \ldots, X_p$ are jointly Gaussian and $X_1 = x_1$ is observed, the new conditional distribution of $X_2, \ldots, X_p | X_1 = x_1$ is also a multivariate Gaussian distribution. In addition, this distribution is completely characterized by its vector of means and its covariance matrix; taken in combination with the previous assumption, we can represent the state of the system at time $t$ by an $n$-dimensional mean vector $\mu_t$ and an $n \times n$ covariance matrix $\Sigma_t$. In particular, the first $n$ observations are drawn from a Gaussian distribution specified by the initial state. Let $Z_t = [Y_{t+1} \cdots Y_{t+n}]^T$ represent the $n$ observations following time $t$. Then

$$Z_0 \sim \mathcal{N}(\mu_0, \Sigma_0).$$

The first assumption raises an implicit question: Given the distributions of the next $n$ observations, how do we extend that to the $(n+1)^{st}$? If we provide an answer to this question, then we can use the current state to compute the distribution for any future observation given the observations seen so far. The answer is given by our *third assumption*: The $(n+1)^{st}$ observation is a weighted sum of the next $n$ observations, plus a Gaussian noise term. However, we do not assume that these noise terms are independent of one another. Formally, at time $t$,

$$Y_{t+n+1} = g^T Z_t + \eta_{t+n+1}, \qquad (5)$$

where $\eta_{t+n+1}$ models the noise in the system and $g$ is the $n$-dimensional vector that defines the linear trend in the sys-tem. The proper way to view (5) is as an extension of the look-ahead horizon—$\mu_t$ and $\Sigma_t$ model the distribution of the next $n$ observations, while (5) allows us to extend the distribution to observations that are *further* in the future. We will see that the distribution of $\eta_{t+n+1}$ becomes important when we observe $y_t$ and wish to update the state.

---

[2]These LDS equations can be extended to the case where the observations, $Y_t$, are each $m$-vectors; we restrict our attention to the case where $m = 1$.

The *fourth assumption* is that the distribution of $\eta_{t+n+1}|h_t$ does not depend upon the value of $t$, where $h_t$ is the history of observations up to time $t$, i.e. $y_1, y_2, \ldots, y_t$. That is,

$$\eta_{t+n+1}|h_t \sim \mathcal{N}(0, \sigma^2). \quad (6)$$

In addition, the covariance of $\eta_{t+n+1}$ with the $n$ observations following $t$ is a constant vector $C$; i.e.

$$\text{Cov}[Z_t, \eta_{t+n+1}|h_t] = C. \quad (7)$$

As mentioned in the second assumption above, because the observations are drawn from a multivariate Gaussian distribution, we capture all information about the distribution of the next $n$ observations with the vector of their means, $\mu_t = \text{E}[Z_t|h_t]$, and their covariance matrix, $\Sigma_t = \text{Var}[Z_t|h_t]$. And because the distribution of the next $n$ observations captures all information about the distribution of *all* future observations, $\mu_t$ and $\Sigma_t$ are a sufficient statistic for $h_t$; that is, $\mu_t$ and $\Sigma_t$ compactly represent all information that is given by observing the system through time $t$.

We can maintain $\mu_t$ and $\Sigma_t$ with straightforward update equations, again because of the Gaussian distribution. Given state $\mu_t$ and $\Sigma_t$, a history $h_t$, and a new observation $Y_{t+1} = y_{t+1}$, we can compute the new state matrices:

$$\mu_{t+1} = G\mu_t + F\frac{y_{t+1} - e_1^T\mu_t}{e_1^T\Sigma_t e_1}, \text{ and} \quad (8)$$

$$\Sigma_{t+1} = G\Sigma_t G^T + B - \frac{FF^T}{e_1^T\Sigma_t e_1}, \quad (9)$$

where $e_i$ is the $i$th column of the $n \times n$ identity matrix,

$$B = \sigma^2 e_n e_n^T + GCe_n^T + e_n C^T G^T,$$

$$F = G\Sigma_t e_1 + C_1 e_n,$$

$$G = \begin{pmatrix} \mathbf{0} & I_{n-1} \\ g^T & \end{pmatrix},$$

$C_1$ is the is the first element of $C$, and $I_{n-1}$ is the $(n-1) \times (n-1)$ identity matrix. See Appendix A.1 for a proof that these equations are the correct state updates; that is, that $\mu_{t+1} = \text{E}[Z_{t+1}|h_t, y_{t+1}]$ and $\Sigma_{t+1} = \text{Var}[Z_{t+1}|h_t, y_{t+1}]$.

**Definition of a PLG** With the update equations (8) and (9), we can define the PLG model. The PLG is the model whose state at time $t$ is $\mu_t$ and $\Sigma_t$ and whose state update equations are (8) and (9). The parameters of a PLG are the initial state ($\mu_0$ and $\Sigma_0$), the linear trend ($g$), and the statistical properties of the noise ($C$ and $\sigma^2$). Note that this is a PSR; the state representation is the mean and variance of future observations (that is, they are predictions about future outcomes given the observations seen so far).

It is instructive to contrast the AR model and the PLG model. The dynamics of the two seem similar (compare (1) and (5)). In fact, if we restrict a PLG such that its noise is i.i.d. (i.e. $C = 0$) and that its covariance matrix is stationary (i.e. $\Sigma_i = \Sigma_j \forall i, j$), they are equivalent. While the PLG may seem simple because it contains no "hidden state," it is in fact as powerful as the LDS model, which does have hidden state. The PLG's representational power comes from the fact that its noise terms covary with the data; this gives it the infinite memory of the LDS—an observation can have an effect far in the future through the chain of influence created by the correlation in the noise terms. However, the similarities between PLGs and AR models are as important as their differences. The fact that PLGs are defined exclusively in terms of observation data, as are AR models, allows us to specify a consistent learning algorithm for PLGs (as we shall see in the next section).

**Theorem 1** *Any LDS with $n$-dimensional state and scalar observations has an equivalent $n$-dimensional PLG.*

We defer the proof of Theorem 1 to Appendix A.3. Note that an $n$-dimensional PLG has fewer parameters than an $n$-dimensional LDS. While $\widehat{x}_1^-$, $P_t^-$, $H$, and $R$ match up to $\mu_0$, $\Sigma_0$, $g$, and $\sigma^2$, $Q$ and $A$ are larger than $C$. An LDS thus has $(3n^2 - n)/2$ more parameters than an equivalent PLG.

The AR and ARMA (Autoregressive Moving-Average) models are also widely used to analyze time-series data. These models are strictly subsumed by LDS models, implying that PLGs strictly subsume them too.

## 3 LEARNING PLGS

Given a corpus of data generated by a dynamical system, we would like to estimate the parameters of a PLG that models that system. Define a *trace* to be a sequence of observations emitted by a dynamical system, starting from the initial state. Then we would like to estimate PLG parameters given $K$ traces, with each trace containing $N$ observations. But how shall we go about this?

When it comes to model learning, PLGs have several potential advantages over LDSs. First, PLGs have fewer parameters than LDSs of the same dimension. In addition, the parameters of PLGs have a definite meaning with respect to the data; for example, a PLG's $\sigma^2$ is the variance of $(Y_{t+n+1} - g^T Z_t)|h_t$. On the other hand, an LDS's $Q$ is the covariance matrix for the latent variables, which is not directly related to the observation data. Finally, there is no inherent symmetry in the PLG parameters; in an LDS, two latent variables can be switched by swapping the ap-propriate rows and columns in $H$, $Q$, $A$, $\widehat{x}_1^-$, and $P_1^-$. This symmetry can cause problems for learning LDS models.

We propose a learning algorithm, which we call the Consistent-Estimation (CE) algorithm, that relies upon the second advantage, the meaning of the PLG parameters. Again, suppose that we are given a dataset composed of $K$ traces, with each trace containing $N$ observations. Let $y_t^k$ be the $t^{th}$ observation from the $k^{th}$ trace.

**Initial State** The initial state matrices, $\mu_0$ and $\Sigma_0$, are just the mean and covariance of the first $n$ observations. Thus, they can be estimated by the sample mean and covariance across all traces of these observations. That is,

$$(\widehat{\mu}_0)_i = \overline{y}_i \equiv \frac{1}{K} \sum_{k=1}^{K} y_i^k \qquad (10)$$

$$(\widehat{\Sigma}_0)_{ij} = \frac{1}{K-1} \sum_{k=1}^{K} (y_i^k - \overline{y}_i)(y_j^k - \overline{y}_j). \qquad (11)$$

Note that these are unbiased estimators for the initial state.

**Linear Trend** The linear trend in the data, $g$, can be estimated by taking advantage of the fact that it *is* the linear trend. That is, $\mathrm{E}[Y_{t+n+1}] = \mathrm{E}[g^T Z_t] + \mathrm{E}[\eta_{t+n+1}] = g^T \mathrm{E}[Z_t]$ (note that these are *not* conditioned on any history). Here we have taken advantage of the fact that $\mathrm{E}[\eta_{t+n+1}] = \mathrm{E}[\mathrm{E}[\eta_{t+n+1}|h_t]]$, by the smoothing property of expectations. Regardless of $h_t$'s value, the inner expectation is 0, and so $\mathrm{E}[\eta_{t+n+1}] = 0$. Thus $\Gamma g = \Lambda$ where

$$\Gamma = \begin{pmatrix} \mathrm{E}[Y_1] & \cdots & \mathrm{E}[Y_n] \\ \vdots & & \vdots \\ \mathrm{E}[Y_{N-n}] & \cdots & \mathrm{E}[Y_{N-1}] \end{pmatrix}, \text{ and}$$

$$\Lambda = (\mathrm{E}[Y_{n+1}] \cdots \mathrm{E}[Y_N])^T.$$

These expectations are not known. However, we can approximate them with sample means of the data. Our estimates of $\Gamma$ and $\Lambda$ given the data are then

$$\widehat{\Gamma} = \begin{pmatrix} \overline{y}_1 & \overline{y}_2 & \cdots & \overline{y}_n \\ \vdots & \vdots & & \vdots \\ \overline{y}_{N-n} & \overline{y}_{N-n+1} & \cdots & \overline{y}_{N-1} \end{pmatrix}, \text{ and}$$

$$\widehat{\Lambda} = (\overline{y}_{n+1} \cdots \overline{y}_N)^T.$$

Substituting these estimates and solving for $g$, we obtain

$$\widehat{g} = (\widehat{\Gamma}^T \widehat{\Gamma})^{-1} \widehat{\Gamma}^T \widehat{\Lambda}. \qquad (12)$$

The noise terms for each row are not independent of each other, so this is in general a biased estimate of $g$. However, as we shall see, its bias shrinks as $K$ increases.

The (unestimated) matrix $\Gamma$ may or may not have full rank. When $\Gamma$ has full rank, there is a unique linear trend, $g$, that can produce the means of the system; we say such systems are UMT (Unique Mean Trend). For non-UMT systems, $\widehat{g}$ may behave poorly as the number of traces increases.

**Noise Parameters** We estimate $C$ and $\sigma^2$ by taking advantage of the fact that they are statistical properties of the noise term, $\eta_{t+n+1}$. Given an estimate of $g$, we can estimate the noise term using its definition in (5); $\widehat{\eta}_{t+n+1}^k = y_{t+n+1}^k - \widehat{g}^T z_t^k$, where $z_t^k$ is the vector $[y_{t+1}^k \cdots y_{t+n}^k]^T$. Then we can estimate $C$ by the sample covariance of the estimated noise with the $n$ preceding observations; $\sigma^2$ can be estimated by the sample variance of the noise estimates. When computing statistics of the noise, we take advantage of the fact that $\mathrm{E}[\eta_{t+n+1}] = 0$ for all $t$:

$$\widehat{C}_i = \frac{1}{K(N-n)-1} \sum_{t=0}^{N-n-1} \sum_{k=1}^{K} y_{t+i}^k \widehat{\eta}_{t+n+1}^k, \text{ and } (13)$$

$$\widehat{\sigma}^2 = \frac{1}{K(N-n)-1} \sum_{t=0}^{N-n-1} \sum_{k=1}^{K} (\widehat{\eta}_{t+n+1}^k)^2. \qquad (14)$$

These estimators are *consistent* whenever the system is UMT: As the number of traces increases, they converge in probability to the true parameters. That is, the probability that they differ from the true parameters vanishes:

**Definition 1** *The sequence $\widehat{x}_1, \widehat{x}_2, \ldots$ converges to $x$ in probability if $\lim_{n \to \infty} \Pr(|\widehat{x}_n - x| > \epsilon) = 0$ for every positive $\epsilon$. We write this as "$\widehat{x}_n \xrightarrow{p} x$ as $n \to \infty$."*

The learning algorithm that computes the estimators defined by (10), (11), (12), (13), and (14) is called the Consistent Estimation (CE) algorithm.

**Theorem 2** *If a dynamical system can be modeled by an $n$-dimensional PLG, is UMT, and generates a training set whose traces are at least $2n$ time steps long, then, as the number of traces $K$ grows, the parameters computed by the CE algorithm will converge in probability to the true parameters of the dynamical system's PLG.*

**Caveats** We prove Theorem 2 in Appendix A.4. This is an important result—a consistent estimation procedure enables us to find a model that is arbitrarily close to the true model, assuming one exists and the training set is large enough. However, it has a weakness: It requires that $\Gamma$ have full rank. This condition could be violated if, for example, $\mu_0 = \mathbf{0}$. Fortunately, this condition is only required to estimate $g$ (see proof); an improved consistent estimator for $g$ would generalize this result to non-UMT systems.

There is an additional caveat to the CE algorithm: while it is consistent and works well with large numbers of traces in its training set, it does not impose constraints on the parameters (i.e. that the variance update of (9) yields a positive semidefinite matrix at each time step). However, this algorithm is the first one suggested by the model. We expect that further development will alleviate this problem.

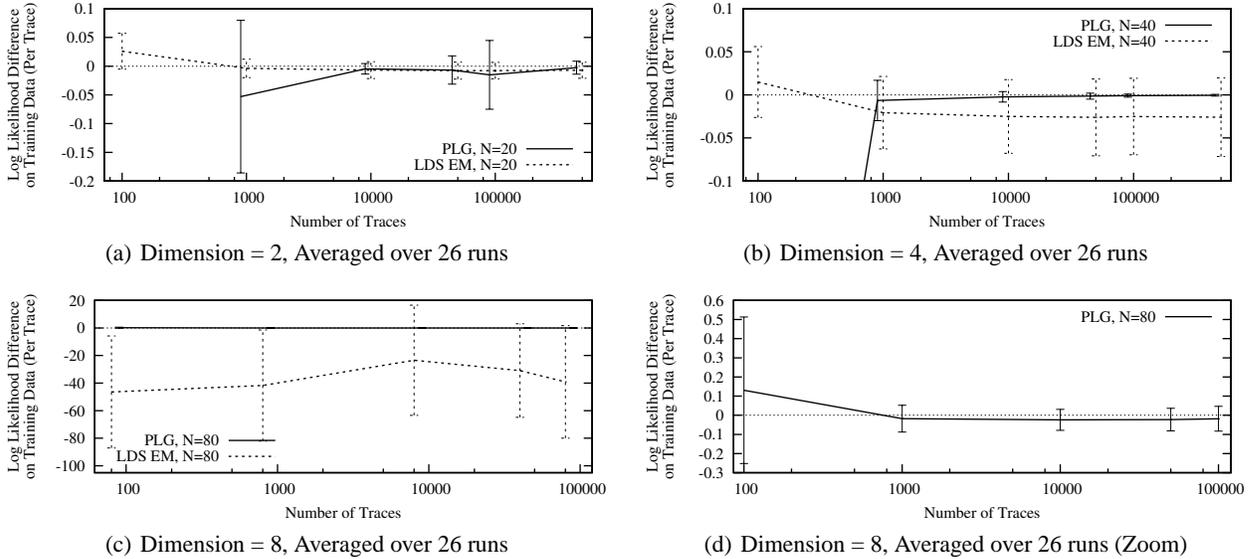

Figure 1: Comparison of EM on LDSs and PLG Learning (Values closer to zero are better)

## 4 EXPERIMENTAL SETUP AND RESULTS

We compare our learning algorithm to the Expectation Maximization (EM) algorithm for LDSs (Ghahramani & Hinton, 1996; Ghahramani, 2002). To compare them, we randomly generated a series of test systems and generated a dataset for each. We then took various subsets from each dataset and trained models using both algorithms. We found that our PLG learning algorithm outperformed EM for large training sets, particularly as the dimensions of the models increased. Additionally, we present some empirical evidence that the parameters learned by our PLG learning algorithm converge toward the true parameters.

**Setup**  We randomly generated LDSs whose dimensions ($n$) were 2, 4, and 8 (Appendix A.5 explains the procedure). To generate a dataset, we first computed the equivalent PLG parameters as in the proof of Theorem 1. We then produced 500,000 traces, each with $10n$ observations, and trained PLGs and LDSs on several subsets of the dataset, running both on 100, 1000, 10,000, 50,000, 100,000, and (except when $n = 8$) 500,000 traces ($K$). For each model learned in this fashion, we reported the error $\frac{l_t - l_a}{K}$, where $l_t$ was the log-likelihood of the training data given the learned parameters and $l_a$ was the log-likelihood of the training data ($K$ traces) given the actual parameters; this is the error in log-likelihood per trace. In each case, the learning algorithms were given $n$—the dimension of the model learned was the same as the dimension of the generating system.

**Results**  The results of these experiments are depicted in Figure 1; each plot shows results averaged over 26 test systems. The number of traces in the training set is plotted on the $x$-axis and the error is plotted on the $y$-axis. Each line is plotted against a slightly perturbed $x$-axis to prevent overlap of error bars; the error bars show one standard deviation in the errors. Values below the dotted line ($y = 0$) show a poor fit of the data. These experiments show that the CE algorithm performs very well compared to EM.

In Figure 1(a), we show the results of the two algorithms when trained on data of dimension 2. Note that the PLG plot starts at $K = 1000$ instead of $K = 100$. This is because 1 of the 26 models learned returned illegal parameters, as described in the previous section. When $n = 2$, the PLG learning algorithm performs comparably to EM when the training set contains 10,000 or more traces, though its variance is greater.

However, when the dimension is 4 (Figure 1(b)), we see that the PLG algorithm begins to outperform EM on a relatively small data set ($K = 1000$); on larger data sets, the PLG had less bias and much smaller variance than EM.

Figures 1(c) and 1(d) show the results when the dimension was 8. Note that the $y$-axis covers a much larger range in Figure 1(c) than in any of the other figures (we would expect the range to be approximately double, as there are twice as many data points per trace). Here, we can see that EM performed relatively poorly. EM's performance never comes close to that of the PLG algorithm, which is shown in more detail in Figure 1(d) by zooming in on the $y$-axis.

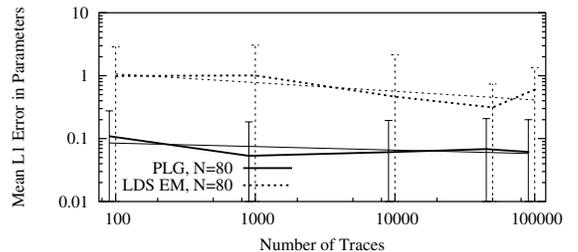

Figure 2: L1 Error in Parameters (Dimension = 8)

We can also see evidence corroborating Theorem 2 in Figure 2 (a log-log plot). This figure shows, for EM and PLG learning, the mean of the L1 norm of the difference between the learned parameters and the true parameter values (divided by the number of parameters) when the dimension was 8. We converted the LDS parameters learned by EM into PLG parameters so that the results are directly comparable. We can see a downward trend (the trendline is plotted in the Figure) in the difference for PLG learning, as the consistency property would lead us to expect.

## 5 CONCLUSION AND FUTURE WORK

We have introduced the Predictive Linear Gaussian model, a predictive model for discrete-time, continuous observation systems. We have shown that the PLG subsumes Linear Dynamical Systems (with scalar observations) and requires fewer parameters. We discussed how the use of statistics over a finite number of future observations as state allows a PLG model to retain the infinite memory property of LDS models without positing unobservable state variables. We have also proposed a consistent learning algorithm for PLG models which takes advantage of the fact that its parameters have statistical meanings with respect to the data. Finally, we have compared this algorithm to Expectation Maximization for LDSs, and shown that our algorithm has some empirical advantages over EM, particularly as the model dimension increases. While EM is not the only (or necessarily the best) algorithm for learning LDSs, it is a widely used algorithm for a well-studied class of models. The fact that CE, an early algorithm for a new class of models, is competitive indicates that CE and PLGs are worth further study.

There are several important areas of future work. First, there are some constraints between the parameters of the PLG; $\Sigma_1$, $g$, $C$, and $\sigma^2$ cannot be specified independently of one another. These constraints must be identified and considered in the learning algorithm. Additionally, maximum likelihood estimation (MLE) should be explored; in cases when few traces are available, our algorithm performs poorly, and MLE may perform better. We are also extending PLG models to allow vector-observations as well as to represent controlled dynamical systems.

Of course, the PLG model cannot completely replace the LDS model. In some applications estimating the value of posited and unobserved state variables is inherently important; for example, a GPS receiver estimates latitude and longitude from satellite signals and reports these estimates to a human user. When this is not the case however, our results indicate that PLG models can be useful because of their potential for more accurate model identification.

## Acknowledgements

The research presented here was supported by a grant from Intel Corp. and by grant IIS 0413004 from NSF.

## A APPENDIX

### A.1 The PLG Update Equations

We show that, if $\mu_t = \mathrm{E}[Z_t|h_t]$ and $\Sigma_t = \mathrm{Var}[Z_t|h_t]$, and if $\mu_{t+1}$ and $\Sigma_{t+1}$ are computed by the state update equations (8) and (9), then $\mu_{t+1} = \mathrm{E}[Z_{t+1}|h_t, y_{t+1}]$ and $\mathrm{Var}_{t+1} = \mathrm{Var}[Z_{t+1}|h_t, y_{t+1}]$. Since $Y_{t+1}, \ldots, Y_{t+n+1}|h_t$ are jointly distributed according to a Gaussian distribution, we can rely on the following lemma:

**Lemma 1** *If the random variable Y and random vector Z are drawn from a multivariate Gaussian distribution with mean $\mu^T = [\mu_Y^T \, \mu_Z^T]$ and covariance matrix*

$$\Sigma = \begin{pmatrix} \sigma_{YY} & \sigma_{YZ}^T \\ \sigma_{YZ} & \Sigma_{ZZ} \end{pmatrix},$$

*then $Z|Y = y \sim \mathcal{N}(\mu_{Z|Y=y}, \Sigma_{Z|Y=y})$, where*

$$\mu_{Z|Y=y} = \mu_Z + \frac{\sigma_{YZ}}{\sigma_{YY}}(y - \mu_Y), \text{ and} \quad (15)$$

$$\Sigma_{Z|Y=y} = \Sigma_{ZZ} - \frac{\sigma_{YZ}\sigma_{YZ}^T}{\sigma_{YY}} \quad (16)$$

*(e.g., Theorem 3.5.2 of Catlin, 1989).*

If we map $Z$ to $Z_{t+1}|h_t$ and $Y$ to $Y_{t+1}|h_t$, we can derive the update equations as follows. The term $\sigma_{YY} = \text{Var}[Y_{t+1}|h_t] = e_1^T \Sigma_t e_1$, which is a scalar (the upper-left element of $\Sigma_t$). Additionally, $\mu_Y = \text{E}[Y_{t+1}|h_t] = e_1^T \mu_t$. Computing $\mu_Z$, $\sigma_{YZ}$, and $\Sigma_{ZZ}$ requires that we compute the distribution of $Y_{t+n+1}|h_t$. From (5) and (6) we have $\text{E}[Y_{t+n+1}|h_t] = \text{E}[g^T Z_t|h_t] + \text{E}[\eta_{t+n+1}|h_t] = g^T \mu_t + 0$, and from (5) and (7), we have

$$\begin{aligned}
\text{Cov}[Y_{t+n+1}, Y_{t+1}|h_t] &= \text{Cov}[g^T Z_t + \eta_{t+n+1}, e_1^T Z_t|h_t] \\
&= g^T \text{Cov}[Z_t, Z_t|h_t] e_1 \\
&\quad + \text{Cov}[\eta_{t+n+1}, Z_t|h_t] e_1 \\
&= g^T \Sigma_t e_1 + C_1,
\end{aligned}$$

where $C_1$ is the first element of $C$. From (5) and (6),

$$\begin{aligned}
\text{Var}[Y_{t+n+1}|h_t] &= \text{Cov}[g^T Z_t + \eta_{t+n+1}, \\
&\quad g^T Z_t + \eta_{t+n+1}|h_t] \\
&= g^T \Sigma_t g + g^T C + C^T g + \sigma^2.
\end{aligned}$$

The mean $\mu_Z = \text{E}[Z_{t+1}|h_t]$; i.e., it is the last $n-1$ elements of $\mu_t$ augmented by $\text{E}[Y_{t+n+1}|h_t]$. That is, $\mu_Z = G\mu_t$. Similarly, $\sigma_{YZ} = G\Sigma_t e_1 + C_1 e_n$. Finally, $\Sigma_{ZZ} = G\Sigma_t G^T + GCe_n^T + e_n C^T G^T + \sigma^2 e_n e_n^T$. Substituting these values into (15) and (16) yields (8) and (9).

### A.2 LDS Prediction with the Kalman Filter

The LDS dynamics are described in (2), (3), and (4). Since $X_t$ is unobservable, we must turn to the Kalman filter, which maintains the state variables $\widehat{x}_t^- = \text{E}[X_t|h_t]$ and $P_t^- = \text{Var}[X_t|h_t]$ (Kalman, 1960; Welch & Bishop, 2004). It maintains these through five equations:

$$\widehat{x}_t^- = A\widehat{x}_{t-1},$$

$$P_t^- = AP_{t-1}A^T + Q,$$

$$K_t = P_t^- H^T (HP_t^- H^T + R)^{-1},$$

$$\widehat{x}_t = \widehat{x}_t^- + K_t(y_t - H\widehat{x}_t^-), \text{ and}$$

$$P_t = (I_n - K_t H)P_t^-.$$

$K_t$ is called the *Kalman gain* at time $t$; $I_n$ is the $n \times n$ identity matrix. In addition,

$$Y_{t+1}|h_t \sim \mathcal{N}(H\widehat{x}_{t+1}^-, HP_{t+1}^- H^T + R). \quad (17)$$

Equation (17) is key—it is through this distribution that the Kalman filter allows us to make predictions about future observations without observing $X_t$.

In particular, an observation $i$ time steps in the future has a Gaussian distribution with the following mean and covariance:

$$\text{E}[Y_{t+i}|h_t] = HA^{i-1}\widehat{x}_{t+1}^-, \text{ and} \quad (18)$$

$$\text{Var}[Y_{t+i}|h_t] = H(A^{i-1}P_{t+1}^-(A^{i-1})^T + S_{i-1})H^T + R, \quad (19)$$

where

$$S_i = \sum_{k=1}^{i} A^{k-1} Q (A^{k-1})^T. \quad (20)$$

In addition, we can compute the covariance of $Y_{t+i}$ and $Y_{t+j}$ where $j \geq i$:

$$\begin{aligned}
\text{Cov}[Y_{t+i}, Y_{t+j}|h_t] &= HA^{j-1}P_{t+1}^-(A^{i-1})^T H^T + \delta_{ij}R \\
&\quad + HA^{j-i}S_{i-1}H^T, \quad (21)
\end{aligned}$$

where $\delta_{ij}$ is the Kronecker delta. We omit this derivation due to lack of space.

### A.3 Proof of Theorem 1

We prove Theorem 1 by construction; given an $n$-dimensional LDS, we compute the parameters of the equivalent PLG and show that the PLG and the LDS compute the same distributions for future observations.

We will show that, given an $n$-dimensional LDS with parameters $A$, $H$, $Q$, $R$, $\widehat{x}_1^-$, and $P_1^-$, there is an equivalent PLG such that, for $n \geq j \geq i \geq 1$,

$$(\mu_0)_i = HA^{i-1}\widehat{x}_1^-, \quad (22)$$

$$\begin{aligned}
(\Sigma_0)_{ij} &= (\Sigma_0)_{ji} \\
&= HA^{j-1}P_1^-(HA^{i-1})^T + \delta_{ij}R \\
&\quad + HA^{j-i}S_{i-1}H^T, \quad (23)
\end{aligned}$$

$$C = \Psi_n - \Psi g - Rg, \quad (24)$$

$$\sigma^2 = HS_n H^T + R - g^T \Psi_n - C^T g, \quad (25)$$

and $g$ is any solution to $g^T M = HA^n$, where $\delta_{ij}$ is the Kronecker delta, $\Psi_i$ is an $n$-vector whose $j^{th}$ element is

$$(\Psi_i)_j = \begin{cases} HA^{i-j+1}S_{j-1}H^T & 1 \leq j \leq i \\ HA^{j-i-1}S_i H^T & i+1 \leq j \leq n \end{cases},$$

$\Psi$ is the $n \times n$ symmetric matrix whose $(i+1)^{th}$ column is $\Psi_i$, $S_k$ is defined by (20), and $M$ is the $n \times n$ matrix whose $i^{th}$ row is $HA^{i-1}$.

The element $(\mu_0)_i$ is $\mathrm{E}[Y_i]$, so (22) follows directly from (18). Likewise, $(\Sigma_0)_{ij}$ is the covariance of $Y_i$ and $Y_j$, so (23) follows from (21).

To compute $g$, note that $\mathrm{E}[Y_{t+n+1}|h_t] = g^T\mu_t = HA^n\widehat{x}_{t+1}^-$ and that $\mu_t = M\widehat{x}_{t+1}^-$ for all $t$—both of these statements follow from (18). Thus $g$ is a solution to $g^TM = HA^n$.

We can compute $C$ by noting that, according to the PLG model, $\mathrm{Cov}[Z_t, Y_{t+n+1}|h_t] = \Sigma_t g + C$; according to (21),

$$\mathrm{Cov}[Z_t, Y_{t+n+1}|h_t] = MP_{t+1}^-(HA^n)^T + \Psi_n.$$

Recall that the elements of $\Sigma_t$ are equal to the covariance computed by (21); i.e., $\Sigma_t = MP_{t+1}^-M^T + \Psi + RI$. Then

$$\begin{aligned}\Sigma_t g &= MP_{t+1}^-M^Tg + \Psi g + Rg \\ &= MP_{t+1}^-(HA^n)^T + \Psi g + Rg.\end{aligned}$$

Thus $C = \Psi_n - \Psi g - Rg$. We can now compute $\sigma^2$:

$$\begin{aligned}\sigma^2 &= \mathrm{Var}[Y_{t+n+1}|h_t] - g^T\Sigma_t g - g^TC - C^Tg \\ &= (HA^nP_{t+1}^-(HA^n)^T + HS_nH^T + R) \\ &\quad - (g^TMP_{t+1}^-(HA^n)^T + g^T\Psi g + Rg^Tg) \\ &\quad - (g^T\Psi_n - g^T\Psi g - Rg^Tg) - C^Tg \\ &= HS_nH^T + R - g^T\Psi_n - C^Tg.\end{aligned}$$

From this derivation of $\sigma^2$ and $C$, it can be seen that, given the correct $\mu_t$ and $\Sigma_t$ and a $g$ that satisfies $g^TM = HA^n$, the PLG model will compute the same distribution for future observations as the Kalman filter for the equivalent LDS. In addition, we have shown that $\mu_0$ and $\Sigma_0$ are the correct initial state variables. Since (15) and (16) govern the conditional distribution of the observations and form the basis for the Kalman filtering equations and the PLG state update, the state variables will remain correct under updating, and we have shown equivalence.

### A.4 Proof of Theorem 2

To prove the consistency of the CE learning algorithm, we will require the following lemma.

**Lemma 2** *If $\widehat{x}_n \xrightarrow{p} x$ as $n \to \infty$, and $f : \mathcal{R}^k \to \mathcal{R}^m$ is continuous at $x$, then $f(\widehat{x}_n) \xrightarrow{p} f(x)$ as $n \to \infty$ (e.g., DeGroot & Schervish, 2002, pg. 234)*

The estimator of the initial mean vector is the sample mean of the first $n$ observations across all traces, and the estimator of the initial covariance matrix is the sample covariance of the first $n$ observations across all traces. These estimators are well known to be consistent.

To show that $\widehat{g}$ is consistent we will require Lemma 2. Thus, we first show that $\widehat{g}$ is continuous in $\overline{y}_1, \overline{y}_2, \dots$ at $\overline{y}_t = \mathrm{E}[Y_t]$. It is clear that $\widehat{\Gamma}$ and $\widehat{\Lambda}$ are everywhere continuous in these variables. We note that any combination of sums, products, and quotients of continuous functions is continuous in the same variables, except where denominators are zero. Therefore, $\widehat{\Gamma}^T\widehat{\Gamma}$ and $\widehat{\Gamma}^T\widehat{\Lambda}$ are everywhere continuous in $\overline{y}_1, \overline{y}_2, \dots$.

If $X$ is nonsingular, $X^{-1}$ can be computed by Gaussian elimination using a combination of sums, products, and quotients of the elements of $X$; division by zero is never necessary. Since $\widehat{\Gamma} = \Gamma$ at $\overline{y}_t = \mathrm{E}[Y_t]$ and $\Gamma$ is full-rank by assumption, $\Gamma^T\Gamma$ is nonsingular and its inverse is continuous in $\overline{y}_1, \overline{y}_2, \dots$ at $\overline{y}_t = \mathrm{E}[Y_t]$. Thus, the product $\widehat{g} = (\widehat{\Gamma}^T\widehat{\Gamma})^{-1}\widehat{\Gamma}^T\widehat{\Lambda}$ is continuous at this point. Since $\overline{y}_t \xrightarrow{p} \mathrm{E}[Y_t]$ (the weak law of large numbers), by Lemma 2 $\widehat{g} \xrightarrow{p} (\Gamma^T\Gamma)^{-1}\Gamma^T\Lambda = g$.

Recall that, by (5), $\eta_{t+n+1}^k = y_{t+n+1}^k - g^Tz_t^k$. Since $\widehat{\eta}_{t+n+1}^k = y_{t+n+1}^k - \widehat{g}^Tz_t^k$ is continuous at $\widehat{g} = g$ and $\widehat{g} \xrightarrow{p} g$, $\widehat{\eta}_{t+n+1}^k \xrightarrow{p} \eta_{t+n+1}^k$. Further, $\widehat{C}_i$ is continuous at $\widehat{\eta}_{t+n+1}^k = \eta_{t+n+1}^k$. As $K \to \infty$,

$$\begin{aligned}\widehat{C}_i &= \frac{1}{K(N-n)-1}\sum_{t=1}^{N-n}\sum_{k=1}^{K} y_{t+i}^k\widehat{\eta}_{t+n+1}^k \\ &\to \frac{1}{N-n}\sum_{t=1}^{N-n}\frac{1}{K}\sum_{k=1}^{K} y_{t+i}^k\widehat{\eta}_{t+n+1}^k \\ &\xrightarrow{p} \frac{1}{N-n}\sum_{t=1}^{N-n}\frac{1}{K}\sum_{k=1}^{K} y_{t+i}^k\eta_{t+n+1}^k \\ &\xrightarrow{p} \frac{1}{N-n}\sum_{t=1}^{N-n}\mathrm{E}[Y_{t+i}\eta_{t+n+1}] \\ &= \frac{1}{N-n}\sum_{t=1}^{N-n}\mathrm{E}[\mathrm{E}[Y_{t+i}\eta_{t+n+1}|h_t]] \\ &= C_i.\end{aligned}$$

The outer expectation in the penultimate line is over all histories $h_t$; the inner expectation is $C_i$ regardless of $h_t$.

By the same argument (*mutatis mutandi*), $\widehat{\sigma}^2 \xrightarrow{p} \sigma^2$.

### A.5 Generating Random LDSs

We randomly generated LDS parameters (refer to (2), (3), and (4) for an explanation of the parameters). Each element of $H$, $A$, and $\widehat{x}_1^-$ was drawn from the uniform distribution $U(-1,1)$. To avoid systems whose observations would tend toward $\pm\infty$, $A$ was normalized so that $\rho(A) = \lambda$ (where $\rho$ denotes the spectral radius and $\lambda \sim U(0,1)$).

We then generated a random correlation matrix, $Q'$, by Marshall and Olkin's (1984) algorithm, and a diagonal matrix $\Sigma$ whose $ii$th element was $2^{x_i}$, with $x_i \sim U(-1,1)$. We computed $Q$ by $\Sigma Q'\Sigma$ (i.e., $Q$ had variances between $1/4$ and $4$ with random correlations). $P_1^-$ was drawn like $Q$, and $R$ was drawn like an element of $\Sigma$.